\documentclass{article} 
\usepackage{iclr2021_conference,times}

\usepackage{amsmath,amsfonts,bm}

\def\eqref#1{equation~\ref{#1}}

\def\1{\bm{1}}

\DeclareMathAlphabet{\mathsfit}{\encodingdefault}{\sfdefault}{m}{sl}
\SetMathAlphabet{\mathsfit}{bold}{\encodingdefault}{\sfdefault}{bx}{n}

\usepackage{hyperref}
\usepackage{url}
\usepackage[utf8]{inputenc}
\usepackage[T1]{fontenc}
\usepackage{comment}
\usepackage{tipa}
\title{Bambara Language Dataset for  Sentiment Analysis}

\author{Mountaga Diallo \\
iCompass \\
49, Street of Marseille, Tunisia \\
\texttt{mount.al.di@gmail.com}
\AND
Chayma Fourati \\
iCompass \\
49, Street of Marseille, Tunisia \\
\texttt{chayma@icompass.digital}
\AND
Hatem Haddad \\
iCompass \\
49, Street of Marseille, Tunisia \\
\texttt{hatem@icompass.digital}
}

\iclrfinalcopy 
\begin{document}

\maketitle
\begin{abstract}

For easier communication, posting, or commenting on each others posts, people use their dialects. In Africa, various languages and dialects exist. However, they are still underrepresented and not fully exploited for analytical studies and research purposes. In order to perform approaches like Machine Learning and Deep Learning, datasets are required. 
One of the African languages is Bambara, used by citizens in different countries. However, no previous work on datasets for this language was performed for Sentiment Analysis. 
In this paper, we present the first common-crawl-based Bambara dialectal dataset dedicated for Sentiment Analysis, available freely for Natural Language Processing research purposes. 

\end{abstract}

\section{Introduction}

Bambara, also called Bamanankan or Bamana, is a language widely used as a vehicular and commercial language in West Africa and one of the national languages of Mali.
Being member of the Mande language family, it is part of the main group in number of speakers, namely the Mandingo language group. This group includes, in addition to Bambara, Dioula in Côte d'Ivoire and Burkina Faso, Mandinka in Senegal and Gambia, as well as the Maninka from Guinea.
According to Worlddata\footnote{https://www.worlddata.info/languages/bambara.php}, Bambara is not an official language in any of these countries, however is spoken as mother tongue by a minor part of the population.
It is most widespread in Mali with a share of around 46\% among citizens. 
For instance, a total of about 15.0 million people worldwide speak Bambara as their mother tongue.


Africa is now experiencing a growth in the use of local languages as means of expression. According to Datareportal\footnote{https://datareportal.com/reports/digital-2021-mali}, there were 2.10 million social media users in Mali in January 2021, who represent 10.2\% of the total population. The number of social media users in Mali increased by 400 thousand, representing an augmentation of 24\% between 2020 and 2021. Most of these people, expressing opinions and ideas online in their own local dialects, are ignored. Hence, the automatic processing of African languages using Natural Language Processing (NLP) becomes crucial. 

In this paper, we introduce the first common-crawl-based Bambara dataset, collected, preprocessed, and annotated for the Sentiment Analysis (SA) task, along with preliminary Machine Learning and Deep Learning experiments' results using our dataset.


\section{Bambara Language}


Bambara is represented as an official language with standard writing rules for various African countries. However, in order to communicate, people express themselves using easier ways of writing. For example, people write in Bambara informally and tend to ignore some of the grammatical and writing rules. Table \ref{examples} presents examples of Bambara sentences with different polarities (positive, negative, and neutral) written informally, and in the official language along with the translation into English .


\begin{table}[htbp]
\caption{Examples of Bambara sentences}
\label{examples}
\begin{center}
\begin{tabular}{ccccc}
\multicolumn{1}{c}{\bf Dialectal Bambara } & \multicolumn{1}{c}{\bf Official Bambara}& \multicolumn{1}{c}{\bf English } &\multicolumn{1}{c}{\bf Polarity }
\\ \hline \\
I togo ?& I t\textopeno g\textopeno ? & What's your name ? & Neutral  \\
I gnanafin ber na & I ny\textepsilon nafin b\textepsilon n’na & I miss you & Positive \\
I Kana maga oun na& I kana magan n na & Don't touch me & Negative \\
\end{tabular}
\end{center}
\end{table}

\section{Related work}



In \cite{tapo2020neural}, a work on neural machine learning translation for extremely low-resourced African languages was performed where a case study on Bambara was conducted.
As a result, the work shows that low-resourced languages present unique challenges to (neural) machine learning translation. 
In this study, most of the resources of parallel texts from Bambara to French or English showed to be low-resourced and hence not practical to use as sources of training data.
The problem was a lack of adherence to the standardized Bambara orthography, due to it being a predominately oral language.



In \cite{mandenkan.2119}, a new online dataset of linguistically rich n-gram frequency data for Bambara is introduced. This dataset is based on the disambiguated part of the Bambara Reference Corpus. These n-grams were constructed with the aim to leverage those types of information that are available in the morphologically annotated corpus of Bambara given the limited amount of textual data. All the corpus is written in official Bambara, which is not the most used one by citizens in everyday communication texts.


In \cite{maslinsky-2014-daba}, a software package entitled Daba was created as part of the corpus development project for Manding languages . Daba has been tested on the Bambara corpus.
The work mentions the lack of standardization of the written form of Bambara and its problem of variability. 

Our works fills this gap by creating the first common-crawl-based Bambara dataset, collected, preprocessed and annotated, which can be used for further research activities in the NLP field.


\section{Data Collection}

\subsection{Bambara V1}
Bambara V1 dataset represents the first version of our common-crawl-based dataset. 
As a result, the dataset contains 2404 sentences written in code-mixed French and Bambara, since people type informally and follow no guidelines such as "mais tu es un grand fan d iba one allah ka kanu to".
In order to have a representative dataset, different topics such as sports, politics, musics, shows, etc. are considered.

Data was preprocessed by removing links, punctuation, and emojis characters. The Sentiment annotation was processed manually by two Malian native speakers who are engineering students. The sentences were annotated as postive (1), negative (-1), or neutral (0).
As a result, the dataset is not balanced including a majority of positive sentences compared to negative and neutral ones.
Table \ref{V1stats} presents Bambara V1 statistics, representing the total number of sentences, number of positive, negative, and neutral sentences, number of words, and number of unique words.

\begin{table}[htbp]
\caption{Bambara V1 Statistics}
\label{V1stats}
\begin{center}
\begin{tabular}{llllll}
\multicolumn{1}{c}{\bf Characteristic}  &\multicolumn{1}{c}{\bf Number }
\\ \hline \\
\#Sentences & 2404 \\
\#Words & 15854 \\
\#Unique words & 3474 \\
\#Positive Sentences & 1587 \\
\#Negative Sentences & 307\\
\#Neutral Sentences & 510 \\

\end{tabular}
\end{center}
\end{table}

\subsection{Bambara V2}
In order to have a balanced dataset, we created the second version of the dataset by collecting new data. 
The same processing and annotation techniques as Bambara V1 were performed. 
As a result, the dataset was enhanced by 497 new sentences. Table \ref{V2stats} presents Bambara V2 statistics including the total number of sentences, number of positive, negative, and neutral sentences, number of words and number of unique words.

\begin{table}[htbp]
\caption{Bambara V2 Statistics}
\label{V2stats}
\begin{center}
\begin{tabular}{llllll}
\multicolumn{1}{c}{\bf Characteristic}  &\multicolumn{1}{c}{\bf Number }
\\ \hline \\
\#Sentences & 3046  \\
\#Words & 19472 \\
\#Unique words & 4244 \\
\#Positive Sentences & 1663 \\
\#Negative Sentences & 579 \\
\#Neutral Sentences & 804\\

\end{tabular}
\end{center}
\end{table}

\section{Experiments and Results}

In this work, we experimented different Machine Learning (ML) and Deep Learning (DL) classifiers with different hyperparameters.
The used models were chosen based on previously best performing ones for SA in the state of the art. 

\subsection{Machine Learning Experiments}
Machine Learning models used are support vector machine (SVM), Random Forest Classifier, Gradiant Boosting Classifier, Logistic Regression and Ridge Classifier. 
The Bambara V1 and Bambara V2 datasets were splitted as Train dataset and Test dataset by ratio 80:20. Different experiments were conducted with a variation of hyperparameters. Metrics used to evaluate the models are accuracy, F1 micro, F1 macro, Recall, and Precision.
Table \ref{results_v1} and table \ref{results_V2} presents the best results obtained for each model using Bambara V1 dataset and Bambara V2 dataset respectively.

\begin{table}[htbp]
\caption{Bambara v1 ML Experiments Results}
\label{results_v1}
\begin{center}
\begin{tabular}{llllll}
\multicolumn{1}{c}{\bf Model}  &\multicolumn{1}{c}{\bf Accuracy}&\multicolumn{1}{c}{\bf F1 Micro }&\multicolumn{1}{c}{\bf F1 Macro}&\multicolumn{1}{c}{\bf Recall}&\multicolumn{1}{c}{\bf Precision}
\\ \hline \\
SVM & \textbf{73\%} & \textbf{58\%} & \textbf{75\%} & \textbf{73\%} & \textbf{77\%} \\
Random Forest Classifier & 71\% & 44\%  & 65\% & 71\% &	65\%  \\
Gradiant Boosting Classifier & 72\% & 47\% &	67\% &	71\% & 66 \%\\
Logistic Regression & 71\% &	42\% &	65\% &	71\% &	66\% \\
Ridge Classifier & 72\%	& 51\% & 69\% & 72\% &	68\% \\
\end{tabular}
\end{center}
\end{table}

\begin{table}[htbp]
\caption{Bambara V2 ML Experiments Results}
\label{results_V2}
\begin{center}
\begin{tabular}{llllll}
\multicolumn{1}{c}{\bf Model}  &\multicolumn{1}{c}{\bf Accuracy}&\multicolumn{1}{c}{\bf F1 Micro }&\multicolumn{1}{c}{\bf F1 Macro}&\multicolumn{1}{c}{\bf Recall}&\multicolumn{1}{c}{\bf Precision}
\\ \hline \\
SVM & \textbf{71\%} & \textbf{65\%} & \textbf{72\% }& \textbf{71\%} &\textbf{ 73\%} \\
Random Forest Classifier & 66\% & 58\% & 64\% & 66\% & 65\% \\
Gradiant Boosting Classifier & 61\% & 49\% & 56\% & 61\% & 61\% \\
Logistic Regression &  68\% & 59\% & 66\% & 68\% & 67\% \\
Ridge Classifier & 71\% & 63\% & 72\% & 71\% & 73\% \\
\end{tabular}
\end{center}
\end{table}

\subsection{Deep Learning Experiments}
The chosen Deep Learning models are two variants of Recurrent Neural Networks (RNN):  Long Short-Term Memory (LSTM) and Bidirectional Long Short-Term Memory (Bi-LSTM).
Since DL models require important amounts of data, the dataset was splitted as train and test with ratio 90:10.
Metrics used to evaluate the models are accuracy, F1 micro, F1 macron Recall, and Precision.  Table 7 and table 8 present the best results obtained for each model using Bambara V1 dataset and Bambara V2 dataset respectively.

\begin{table}[htbp]
\caption{Bambara v1 DL Experiments Results}
\label{results_dp_v1}
\begin{center}
\begin{tabular}{llllll}
\multicolumn{1}{c}{\bf Model}  &\multicolumn{1}{c}{\bf Accuracy}&\multicolumn{1}{c}{\bf F1 Micro }&\multicolumn{1}{c}{\bf F1 Macro}&\multicolumn{1}{c}{\bf Recall}&\multicolumn{1}{c}{\bf Precision}
\\ \hline \\
LSTM & \textbf{77\%} & \textbf{77\%} & \textbf{64\% }& \textbf{77\%} &\textbf{ 77\%} \\
Bi-LSTM & 72\% & 72\% & 60\% & 72\% & 72\% \\
\end{tabular}
\end{center}
\end{table}

\begin{table}[htbp]
\caption{Bambara V2 DL Experiments Results}
\label{results_dp_V2}
\begin{center}
\begin{tabular}{llllll}
\multicolumn{1}{c}{\bf Model}  &\multicolumn{1}{c}{\bf Accuracy}&\multicolumn{1}{c}{\bf F1 Micro }&\multicolumn{1}{c}{\bf F1 Macro}&\multicolumn{1}{c}{\bf Recall}&\multicolumn{1}{c}{\bf Precision}
\\ \hline \\
LSTM & \textbf{69\%} & \textbf{69\%} & \textbf{64\% }& \textbf{69\%} &\textbf{ 69\%} \\
Bi-LSTM & 69\% & 69\% & 63\% & 69\% & 69\% \\
\end{tabular}
\end{center}
\end{table}

\section{Discussion}
As a result, fairly satisfactory results were obtained. After training different ML models, the best performing model proved to be Support Vector Machine (SVM) for both versions of the dataset.
This could be explained by the particularity of SVM to separate classes. SVM is the only linear model which can classify data that is not linearly separable and allows a good generalization of unseen data samples.

For the Deep Learning experiments, the obtained LSTM and Bi-LSTM results are quite close. LSTM outperforms Bi-LSTM for the two versions of the dataset.  
This could be explained by the LSTM’s ability to forget, remember and update the information during training.

The first version of the dataset (Bambara V1) performed better results using the DL approach. However, results proved that the best performing ML model (SVM) achieves better results than the Deep Learning one (LSTM) for the Bambara V2 dataset. This could be explained by the relatively small size of the dataset, where DL models require large amounts of datasets for training.

\section{Conclusion and Future Work}

In this paper, we introduced the first Bambara dataset collected, preprocessed, and annotated for Sentiment Analysis, along with experiments performed using ML and DL models. The results obtained are quite satisfactory.
The datasets are publicly available\footnote{https://github.com/chaymafourati/BAMBARA-LANGUAGE-DATASET-FOR-SENTIMENT-ANALYSIS} (Open Data Sources) for further research activities and analytical studies.
Because of some African dialects similarities, our dataset can also be used for training other Mande languages such as Dioula or Maninka.
As a future work, we plan to increase the size of Bambara V2 dataset and build other underrepresented  African dialects in order to help the African NLP community for further research activities.

\bibliography{iclr2021_conference}
\bibliographystyle{iclr2021_conference}
\end{document}